# Concepts is All You Need: A More Direct Path to AGI

Peter Voss, AIGO.ai, Austin TX, USA, peter@aigo.ai

Mlađan Jovanović, Singidunum University, Belgrade Serbia, mjovanovic@singidunum.ac.rs

**Abstract** – Little demonstrable progress has been made toward AGI (Artificial General Intelligence) since the term was coined some 20 years ago. In spite of the fantastic breakthroughs in Statistical AI such as AlphaZero, ChatGPT, and Stable Diffusion none of these projects have, or claim to have, a clear path to AGI. In order to expedite the development of AGI it is crucial to understand and identify the core requirements of human-like intelligence as it pertains to AGI. From that one can distill which particular development steps are necessary to achieve AGI, and which are a distraction. Such analysis highlights the need for a Cognitive AI approach rather than the currently favored statistical and generative efforts. More specifically it identifies the central role of concepts in human-like cognition. Here we outline an architecture and development plan, together with some preliminary results, that offers a much more direct path to full Human-Level AI (HLAI)/ AGI.

## Requirements of General Intelligence

We expect an AGI to be capable of performing any cognitive task, especially novel ones, at a level comparable to a human being [1]. Thus, a key feature of general intelligence is the ability to learn new knowledge and skills. As far as AGI design is concerned, it is much more important to be able to *acquire* knowledge than simply *having* it.

Moreover, AGI has core requirements as to *what* to learn, and *how* to learn as follows:

- Be able to learn real-world 4D data that could be noisy, incomplete, or even wrong.
- Such data includes new entities and action sequences, plus their relationships.
- Knowledge must be interpreted, encoded, and evaluated *conceptually*. This means that entities, actions, and generalizations are represented by their (scalar) attributes. In other words, as vectors with a schema. This is essential to facilitate (dis)similarity comparisons, and for forming higher-level abstract concepts.
- The ability to learn complex data such as images and movement *interactively*. This implies having input senses and output actuators of some kind, plus mechanisms to select and extract particular input data (selective attention) [2].
- The system must be able to accumulate new knowledge and skills *incrementally*, integrating with existing *short- and long-term memory*. This requires a robust knowledge representation such as an integrated, high-performance knowledge graph.
- Input needs to be interpreted *contextually*, taking into account prior input and knowledge as well as current goals and priorities.
- Learning must be *life-long* and *adaptive* with the ability to change or invalidate existing knowledge and to adjust to new situations and environments.

- Most learning should be **autonomous** (unsupervised or self-supervised), without a human in the loop.
- The system must operate in **real-time**, and function adequately with limited resources [3].
- Human-like intelligence covers a wide range of **learning modes**, including: instance or one-shot; clustering and association by time and/or space; generalization; aping; stimulus-response; reinforcement; random and structured exploration; human guided; via instructions; as well as zero-shot (implicit inference); explicit reasoning (figuring things out) and study (read, view).

Effective AGI designs must not only implement all of these learning requirements, but also embody methods for action control, reasoning mechanisms, and metacognitive control.

## Cognitive AI

Essential requirements of AGI cannot be met by logic or statistical methods alone [4] – they demand a cognitive approach. A DARPA presentation makes a useful distinction with 'The Three Waves of AI' [5]:

- 1st Wave: Rule-based, hand-crafted knowledge – GOFAI, expert systems, Deep Blue.
- 2nd Wave: Statistical learning – DL/ML/RL/Transformers, AlphaZero, ChatGPT.
- 3rd Wave: Cognitive AI, autonomous contextual adaptation – Cognitive architectures.

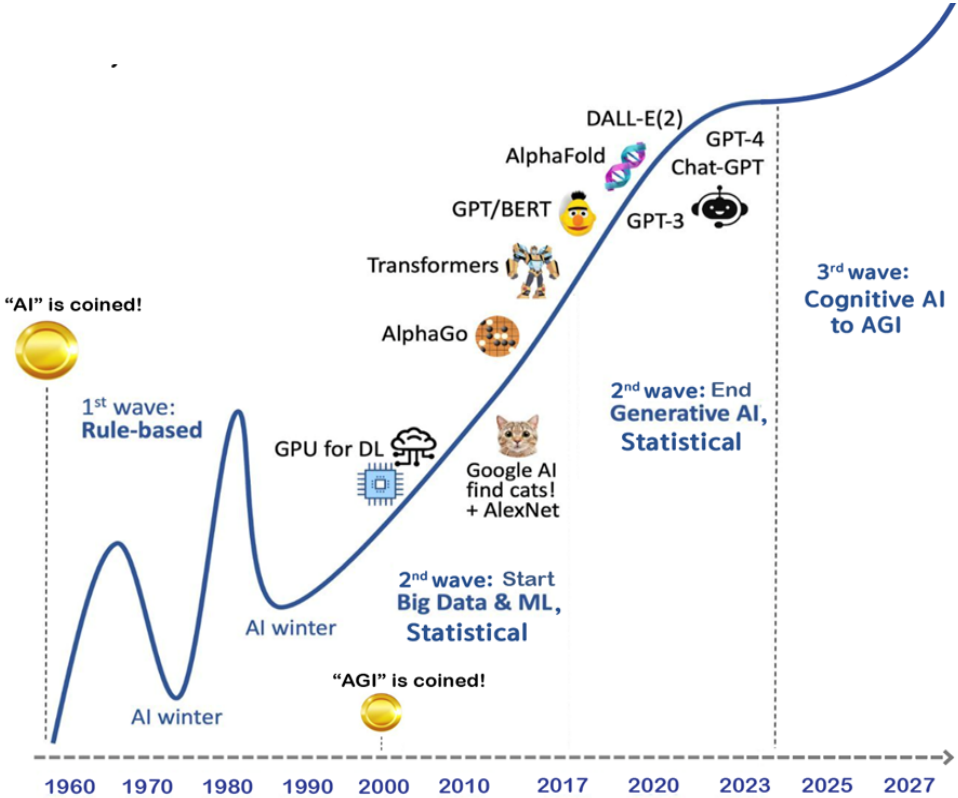

**Figure 1.** Timeline of growing capabilities of AI systems.

The Cognitive AI approach is *not* just a mashup of the first two 'Waves', though it freely incorporates insights gained from those earlier methods. It is typically implemented as a cognitive architecture. We describe Cognitive Architectures as systems that encompass and embody all of the essential structures required for a human-level mind [1]. It also considers how these structures and functions need to work together effectively and function intelligently in diverse, dynamic environments [6].

## Senses and Actuators

We perceive objects and actions via our senses. Subconscious, lower-level processes 'package' input data streams into digestible objects that we become aware of. One could argue that an AGI needs to fully integrate such preprocessing with higher-level cognition. This view certainly has merit; higher-level context influences lower-level focus and selection, and recognition. However, both theoretical consideration as well as practical experience indicate that effective AGI can be constructed with separate pre-processing mechanisms for visual, tactile, and sound input [7].

An insightful perspective to consider is that both Helen Keller, with severely limited sense perception, and Stephen Hawking, with little dexterity, were able to deliver outstanding intellectual contributions. We could call this the 'Helen Hawking' model of AGI – a powerful cognitive system with very limited sense acuity and dexterity.

From a practical point of view, we posit that a highly effective AGI could be limited to something like PC desktop visual input supplemented by text or sound, plus the ability to manipulate mouse and keyboard. Screen input would potentially provide a real-time window to the real world. A limiting factor may be the lack of direct 3D or depth perception, that blind people obviously obtain via touch and sound.

## From Percepts to Concepts

Returning to the nature of what we (and potential AGIs) perceive, it is crucial to note that the objects and actions are essentially vectors composed of numerous scalar features. All knowledge, actions, and skills that we learn and utilize cognitively can be seen as vectors plus their relevant relationships.

For example, facial features can be represented by a number of features – whether via simple, traditional low-dimensional distance measures, or via modern complex machine learning features [8]. Similarly, actions are represented by various scalar dimensions – e.g., a ball bouncing, rolling, or floating with values for frequency, amplitude, speed, etc. [9].

Vector representation forms the basis not only for lower-level cognitive functions such as similarity measures, but also ultimately for our uniquely human ability to form highly abstract concepts, and to be able to reason with them.

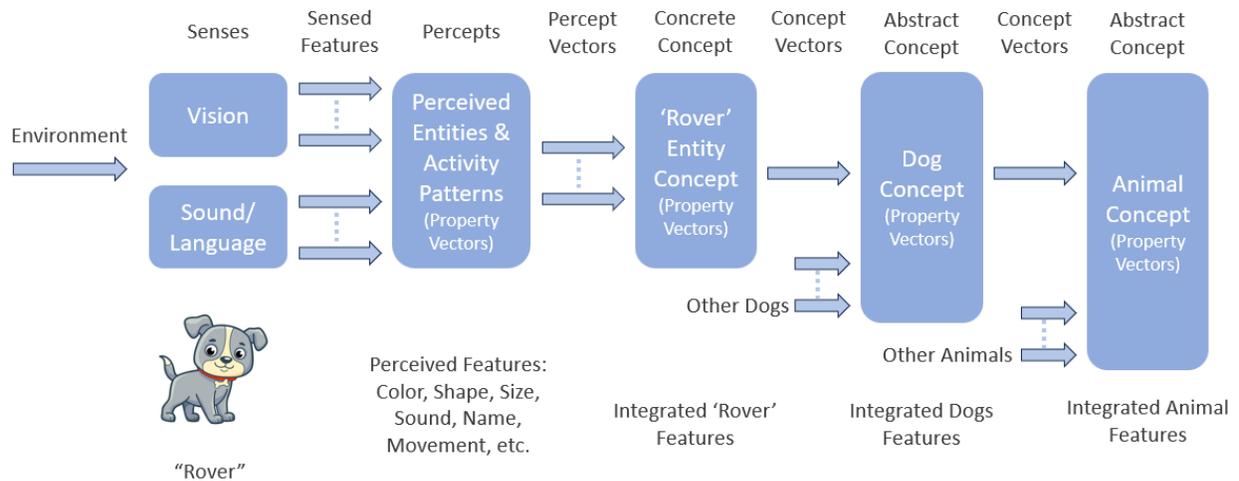

**Figure 2.** Entity and activity features from senses are stored or recognized as percept vectors. These in turn are integrated into entity ('Rover'), concept (dog), and abstract category levels (animal).

The utility of vector encoding **and hierarchies** is amply demonstrated by the power of LLMs [10]. However, this Statistical AI approach suffers three important limitations:

1. These vectors are based on co-occurrence or prediction-relevance in training data, and not on real-world ontological features.
2. They have fixed dimensionality rather than one most appropriate to each concept. These limitations ultimately hamper robust ongoing learning and reasoning.
3. Vectors in LLMs are established during training and do not change during 'inference' – while interacting with users.

The proposed Cognitive AI approach does not inherently suffer from these issues. It provides for variable-size vectors that are dynamically adaptive, and are more directly grounded with real-world features.

## Knowledge Representation

Core AGI requirements dictate the need for a long-term memory store of vectors representing things like entities, concepts, action sequences, and various relationships. Graph-like data stores are ideally suited for this purpose. They provide the flexibility of encoding a myriad of different complex structures and relationships.

However, performance considerations rule out the use of external graph databases because all recognition, learning, and cognitive functions need to constantly reference and update the graph. Only a custom, fully integrated, memory-based knowledge-graph system can provide the speed required to operate in real-time. Recent benchmark tests have shown a 1000-fold difference in access time between these two approaches (Table 1).

**Table 1.** Given a data sample for a Graph DB, the table illustrates how quickly AIGO KG and Neo4j Graph DB can find a node (contact the authors for the experiment details).

| Calls/searches | AIGO KG | Neo4j Graph DB | AIGO faster by |
|---|---|---|---|
| 1 | ~ 0 ms | 6 ms | x→∞ |
| 10 | ~ 0 ms | 11 ms | x→∞ |
| 100 | ~ 0 ms | 83 ms | x→∞ |
| 1.000 | 1 ms | 838 ms | ~838x |
| 100.000 | 49 ms | 73,809 ms | ~1500x |
| 1.000.000 | 446 ms | 747,017 ms | ~1670x |

Traditionally, cognitive architectures have been implemented in a very modular fashion which, generally speaking, is good engineering practice [11]. For AGI, however, we need extremely tight integration between the various cognitive mechanisms. Context, memory, pattern matching, learning, generalization, inference, exploration, action and metacognition constantly interact in complex ways.

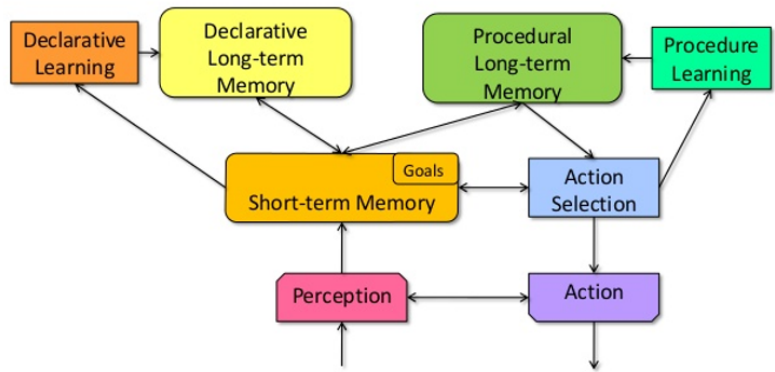

**Figure 3.** Traditional, Modular Cognitive Architecture Design.

A powerful way to achieve this is to have a hyper-optimized graph-based vector datastore act as a foundational substrate for all cognitive functions. Not only does this vector graph serve as long-term memory, but it can also double as short-term memory via suitable activation mechanisms. Such a system has to be carefully designed and built from the ground up to ensure full integration and good performance. Off-the-shelf components or separate modules won't do.

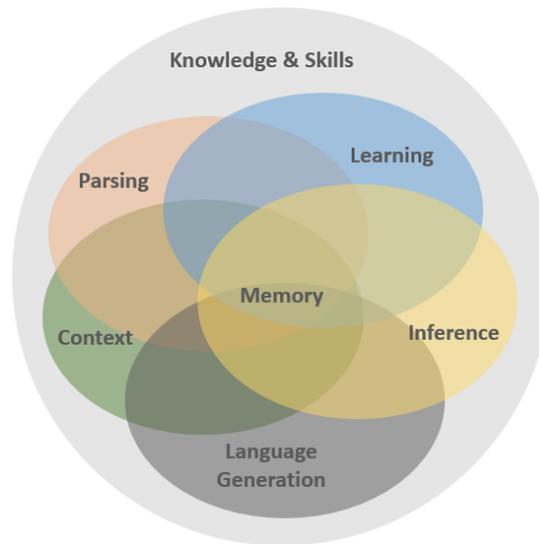

**Figure 4.** Outline of Fully Integrated Cognitive Architecture using a Knowledge-Base Substrate.

## Metacognition and Emotions

The majority of our cognition is subconscious – it does not involve explicit mental control or supervision. However, what sets human intelligence apart is that we are able to think about, and direct our thinking. This distinction was well articulated by Kahneman's System1/System2 model [12]. It is important to note that the distinction is not binary, but rather transitions from one to the other seamlessly.

Higher-level, or System 2, cognition is one aspect of overall awareness and control. The other pertains to the availability of what one could call 'cognitive emotions', mental states such as surprise, certainty, confusion, and boredom [13]. These signals affect both conscious as well as subconscious cognition. They can also be controlled to some extent; much more so in AIs than humans.

Both of these powerful mechanisms need to be an integral part of any workable AGI design.

## AGI Curriculum and Benchmarks

One of the fundamental principles of Cognitive AI is that it should be built with a minimum of hard-coded or fixed functionality. It should be as adaptive as possible, both as far as new knowledge and skills are concerned, as well as being able to adapt to new information and circumstances.

Additionally, its hierarchical, conceptual knowledge base should be as accurate and grounded as possible. These requirements combine to put a large burden on the *quality* of training – especially for its foundational knowledge. Unlike LLMs, Cognitive AI doesn't inherently need massive amounts of uncurated training data, instead it needs a carefully designed curriculum to create a robust hierarchy of knowledge and skills.

A related difficulty is the design of tests and benchmarks. Existing SOTA benchmarks are not appropriate for evaluating early-stage AGI designs [14]; an early general AI would not be expected to do well on specialized narrow tasks, or on tasks that require a wide range of knowledge.

Performance tests can also not be too generic, they have to be designed to measure progress in relation to the specific AGI theory involved [15], to the types of sense and activators used, as well as the chosen curriculum.

To minimize the risk of designing tests aligned with what the system *can* do, rather than what it ***should*** be able to do, these benchmarks should ideally be developed by an external party that does however have a very good understanding of the overall setup and theory (see 'Benchmarks for Proto-AGI', in preparation at the time of writing this article).

## Practical Implementations

The 'Aigo' project (originally, 'a2i2' Adaptive A.I Inc) has over the past 20 years produced a number of AGI development prototypes (as well as several 'industrial grade' commercial versions) using this approach[1]. All of these systems utilize a graph-based knowledge substrate into which all cognitive subsystems are integrated. Early models incorporated several sense inputs (vision, sound, touch, etc.) operating in a simulated environment, while later commercial versions focused on speech and text IO (input/output). These conversation-focused implementations demonstrate powerful real-time contextual learning, memory capabilities, as well as reasoning and question answering[2].

One of our benchmarks (conducted at the end of August 2023) was to test the ability to learn novel facts and answer questions about them. We compared AIGO with Chat GPT-4 (8.000 tokens context window) and with Claude 2 (100.000 tokens context length). The AIGO system was pretrained with only a rudimentary real-world ontology of a few thousand general concepts such as person, animal, red, and small. Chat GPT-4 and Claude 2 were used in their standard form and not constrained in any way.

The test involved first feeding 419 natural language statements to each of the three systems. These were simple facts, *some* of which related to each other (e.g., Tina wants a dog and a cat. Actually, Tina only wants a cat). Finally, we asked 737 questions and scored the answers. We evaluated the responses based on a reasonable human standard. If the response pertains to the topic, answers correctly based on the correct source of information, and is grammatically sound, we consider the answer correct.

The AIGO system scored 88.89%, whereas Claude 2 only managed 35.33%. Chat GPT-4 was unable to perform the test. It scored less than 1%. Details and analysis of the experiment are available elsewhere[3] and from the authors.

## A Roadmap to AGI

The current Aigo baseline system offers excellent knowledge-graph performance, deep contextual parsing and understanding, real-time adaptive learning, and integrated inference. It does however no longer support multi-modal input or output, or have low-level, integrated vector support.

A recently launched Aigo development project revisits multi-model IO and aims to eliminate various existing rule-systems, and to significantly reduce the amount of code. The curriculum and assessments are specially crafted to foster the system's increasing autonomy in knowledge and skill acquisition. The vast range of currently available LLMs, along with data resources such as Wikipedia, will significantly aid in the development of curated knowledge acquisition.

---

[1] Aigo.ai Web Chat Demo: https://www.youtube.com/watch?v=FLQPdS9tvQg (Accessed 19.08.2023).
[2] Aigo.ai Elder Companion Demo: https://www.youtube.com/watch?v=VVumDGlSRng (Accessed 17.08.2023).
[3] Aigo.ai Benchmark: http://tinyurl.com/2x59ma4d (Accessed 31.08.2023).

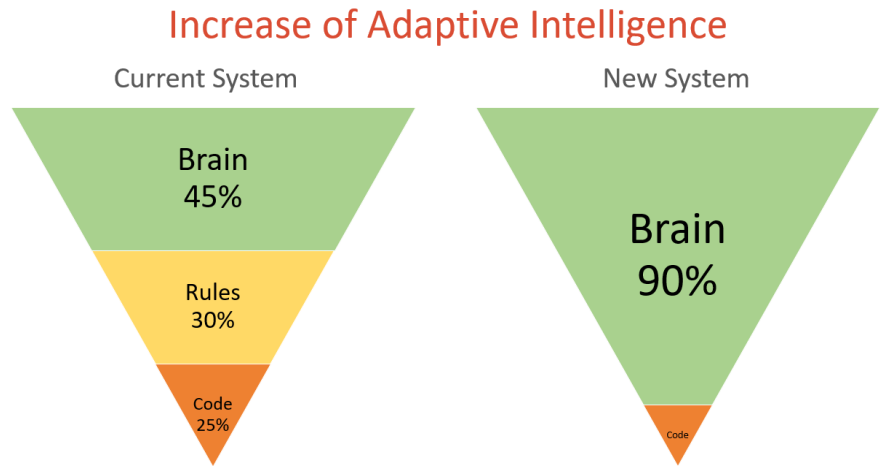

**Figure 5.** The current architecture's rules systems, as well as various functions currently hard-coded, need to be re-implemented as concept structures ('Brain'). It will more fully integrate this functionality with overall cognition, and also make it fully adaptive.

A roadmap outline to upgrading the system to fully meet the core requirements of AGI includes the following activities:

- Re-integrating multimodal vector pattern learning and matching into the knowledge graph.
- Adding real-time and background abstraction/concept formation.
- Training system to do basic question-answering.
- Training and developing advanced language capabilities.
- Adding multi-modal action and action learning mechanisms.
- Training semi-autonomous incremental knowledge acquisition and validating using multiple sources.

At this point the system will have basic human-level (HLAI) 'High School' capabilities. Further iterative enhancements include:

- Semi-autonomous acquisition of conversation requirements (language-based assistant).
- Developing advanced visual sensing and output (mouse) dexterity.
- Advanced space and time representation and modeling.
- Extending metacognition, focus-and-selection, and other control mechanisms.
- Advanced multi-modal learning, reasoning and problem solving (PC-based assistant).
- Significantly scaling up the amount of embedded common knowledge and high-level reasoning skills.

This will bring the system to 'Graduate' level. Ongoing improvements in autonomous learning, high-level reasoning (including Theory-of-Mind) as well as capacity and performance enhancements will bring Aigo up to full AGI capabilities.

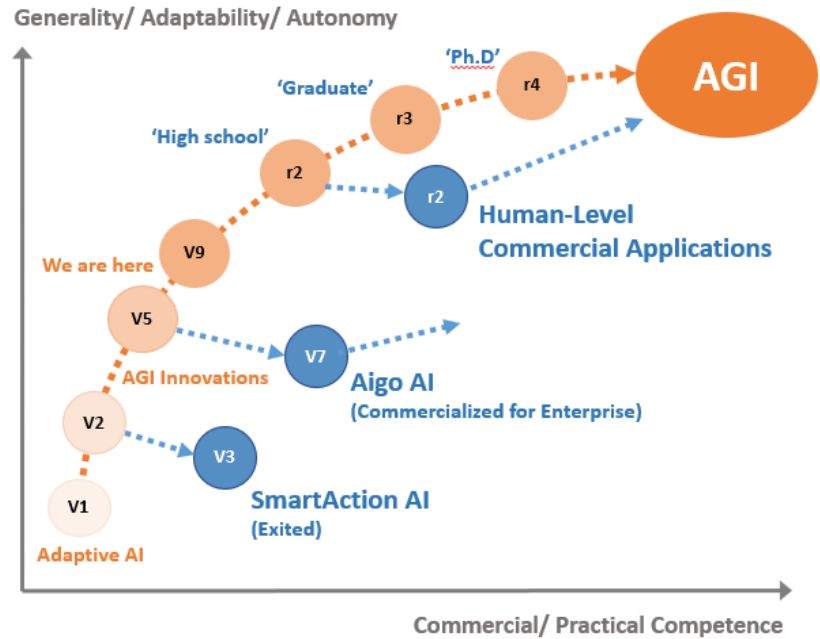

**Figure 6.** Past, present and future Aigo roadmap milestones.

## Conclusion

Progress towards AGI has been much slower than expected or necessary. A key reason for this is the lack of focus on what human-like cognition really requires, thus missing key properties of high-level intelligence. Crucial features, such as autonomous, incremental, real-time learning and adaptation, cannot be adequately addressed by Statistical AI; they require a Cognitive AI approach. We detail some of these often-overlooked features, and specifically highlight the need for effective *conceptual* knowledge representation. We introduce 'Aigo', a high-performance, highly integrated cognitive architecture that has over the past 20 years been utilized both for AGI research, as well as for advanced commercial 'Conversational AI' applications. This architecture is being extended to meet all of the core requirements of AGI in order to achieve human-level adaptive autonomous intelligence.

10